\documentclass[10pt,twocolumn,letterpaper]{article}

\usepackage{cvpr}              

\usepackage{graphicx}
\usepackage{booktabs}
\usepackage{amsmath}
\usepackage{amssymb}
\usepackage{booktabs}
\usepackage{multirow}
\usepackage{pifont}
\usepackage{enumitem}
\usepackage{marvosym}
\usepackage{xcolor}

\newcommand{\cmark}{\ding{51}}%
\newcommand{\xmark}{\ding{55}}%
\definecolor{cvprblue}{rgb}{0.21,0.49,0.74}
\usepackage[pagebackref,breaklinks,colorlinks,allcolors=cvprblue]{hyperref}  
\PassOptionsToPackage{bookmarks=false}{hyperref}

\usepackage[capitalize]{cleveref}
\crefname{section}{Sec.}{Secs.}
\Crefname{section}{Section}{Sections}
\Crefname{table}{Table}{Tables}
\crefname{table}{Tab.}{Tabs.}

\def\cvprPaperID{9150} 
\def\confName{CVPR}
\def\confYear{2025}

\begin{document}

\title{Grounding 3D Object Affordance with Language Instructions, Visual Observations and Interactions}

\newcommand*{\affaddr}[1]{#1} 
\newcommand*{\affmark}[1][*]{\textsuperscript{#1}}
\newcommand*{\email}[1]{\texttt{#1}}
\author{
He Zhu$^{1, 2}$ \quad Quyu Kong$^2$ \quad Kechun Xu$^{1,2}$ \quad Xunlong Xia$^2$ \quad Bing Deng$^2$ \\ Jieping Ye$^2$ \quad Rong Xiong$^{1}$\thanks{Corresponding author (rxiong@zju.edu.cn).} \quad Yue Wang$^{1}$ \\
\affaddr{\affmark[1]Zhejiang University} \quad \affaddr{\affmark[2]Alibaba Cloud} \\
}
\maketitle
\begin{abstract}
   Grounding 3D object affordance is a task that locates objects in 3D space where they can be manipulated, which links perception and action for embodied intelligence. For example, for an intelligent robot, it is necessary to accurately ground the affordance of an object and grasp it according to human instructions. In this paper, we introduce a novel task that grounds 3D object affordance based on language instructions, visual observations and interactions, which is inspired by cognitive science. We collect an Affordance Grounding dataset with Points, Images and Language instructions (AGPIL) to support the proposed task. In the 3D physical world, due to observation orientation, object rotation, or spatial occlusion, we can only get a partial observation of the object. So this dataset includes affordance estimations of objects from full-view, partial-view, and rotation-view perspectives. To accomplish this task, we propose LMAffordance3D, the first multi-modal, language-guided 3D affordance grounding network, which applies a vision-language model to fuse 2D and 3D spatial features with semantic features. Comprehensive experiments on AGPIL demonstrate the effectiveness and superiority of our method on this task, even in unseen experimental settings. Our project is available at  \url{https://sites.google.com/view/lmaffordance3d}.
\end{abstract}

\section{Introduction}
\label{introduction}
Affordance is the property of an object that suggest how it can be used\cite{gibson1977theory}. Grounding 3D object affordance aims to classify, detect, or segment the object's affordances in a scene with the input of images, videos, and point clouds, which plays an important role in robotics, human-object interactions (HOI), scene
understanding, functionality understanding, and other fields\cite{chen2023survey, hong20233d}.

\begin{figure}[t]
  \centering
   \includegraphics[width=0.825\linewidth]{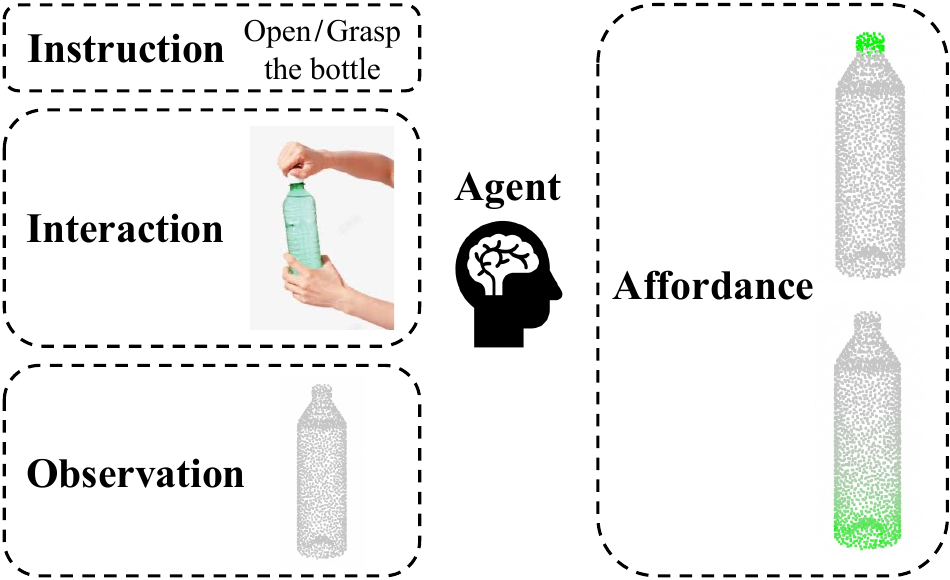}
   \caption{\textbf{Illustration for the affordance grounding task.} Inspired by cognitive science, when humans encounter a new object, they learn its affordance through language instructions, vision information from eyes, and human-machine interactions, thus obtaining its affordance.}
   \label{illustration}
   \vspace{-0.36cm}
\end{figure}

Although there are numerous studies\cite{do2018affordancenet, fang2018demo2vec, nagarajan2019grounded, luo2022learning} focusing on grounding object affordances from 2D images or videos, the knowledge of affordances derived from 2D data remains challenging to apply in 3D space.
Some works\cite{deng20213d, xu2022partafford} have been dedicated to using 3D data to ground object affordances. They learn from geometry to segment out affordance regions from point clouds. However, they exhibit limited generalization for grounding unseen object affordances and tend to show confusion when encountering similar objects.\cite{yang2023grounding}. Actually, humans learn the affordances of a new object from demonstrations in images or videos, as well as from verbal instructions. It is a complex and comprehensive task that involves understanding of language instructions, geometric relationships, and scene information. Cognitive science research\cite{hartson2018ux} also indicates that visual information and auditory language cues humans perceive can enhance their ability to use or manipulate new objects. Inspired by this, we introduce a novel task that grounds 3D object affordance based on language instructions, visual observations, and interactions.

\begin{table*}[tb]
  \centering
  \renewcommand{\arraystretch}{0.8} 
  \resizebox{\textwidth}{!}{%
  \begin{tabular}{lccccccc}
    \toprule
        \multirow{2}{*}{\textbf{Dataset}} & \multirow{2}{*}{\textbf{Year}} & \multicolumn{3}{c}{\textbf{Input}} & \multirow{2}{*}{\textbf{Output}} & \multirow{2}{*}{\textbf{View}} & \multirow{2}{*}{\textbf{Number}} \\ \cmidrule(r){3-5} && \textbf{Image} & \textbf{Point cloud} & \textbf{Language} && \\
        \midrule
        PAD\cite{luo2021one} & 2021 & \cmark & \xmark & \xmark & 2D Segmentation & - & 4002 \\
        CAD-120++\cite{han2022one} & 2022 & \cmark & \xmark & \xmark & 2D Box & - & 32327 \\
        AGD20K\cite{luo2022learning} & 2022 & \cmark & \xmark & \xmark & 2D Heatmap & - & 20061 \\
        3D AffordanceNet\cite{deng20213d} & 2021 & \xmark & \cmark & \xmark & 3D Heatmap & F, P, R & 56307\\
        GAPartNet\cite{geng2023gapartnet} & 2023 & \xmark & \cmark & \xmark & 3D Segmentation \& Pose & P, R & 8489 \\
        PIAD\cite{yang2023grounding} & 2023 & \cmark & \cmark & \xmark & 3D Heatmap & F & 7012\\
        LASO\cite{li2024laso} & 2024 & \xmark & \cmark & \cmark & 3D Heatmap & F & 19751 \\
        AGPIL (Ours) & 2024 & \cmark & \cmark & \cmark & 3D Heatmap & F, P, R & 30972\\
    \bottomrule
    \end{tabular}
    }
    \caption{\textbf{Comparion of affordance grounding dataset.} Here we mainly compare the input, output, view, and dataset scale for different datasets. F, P, and R respectively refer to full-view, partial-view, and rotation-view.}
    \label{Dataset_comparion}
    \vspace{-0.3cm}
\end{table*}

In this paper, we present AGPIL, which is the first multi-modal and multi-view dataset for grounding 3D object affordance. For the full-view dataset, each frame of data includes a language instruction, an image of human-object interactions, and point clouds of the object's entire external surface. In the 3D physical world, due to observation orientation, object rotation, or spatial occlusion, we can only acquire partial scans of the object. So we also build the partial-view and rotation-view datasets. The partial-view dataset includes only one side or part of the object's point clouds, while the rotation-view dataset contains part of the object’s point clouds with random rotation along the x, y, and z axes. In addition, each view of datasets has two settings: seen and unseen, which can comprehensively test the model's generalization performance.

To address the task of grounding 3D object affordance, we propose LMAffordance3D, which is an end-to-end multi-modal, language-guided 3D affordance grounding framework based on the vision-language model (VLM). And This framework operates as a one-stage method. The key idea is to integrate 2D and 3D spatial features and project them into a text space via multilayer perceptron (MLP). We then concatenate all the features as input for the vision-language model. Using a decoder, we fuse the output from the vision-language model with the 2D and 3D spatial features, subsequently employing a segmentation head to delineate affordance regions.

In summary, our contributions are mainly two-fold:
\begin{itemize}
\item We introduce a novel task that grounds 3D object affordance based on language instructions, visual observations, and interactions. And we present AGPIL which is composed of 30972 point-image-text pairs, covering 17 affordance categories and 23 object classes. It has full-view, partial-view, and rotation-view in seen and unseen settings. To the best of our knowledge, this is the first multi-modal and multi-view 3D affordance dataset with well-defined probabilistic affordance score annotations.
\item We establish a comprehensive benchmark based on AGPIL for three different views in two settings. And we propose LMAffordance3D as a baseline, which is a novel end-to-end multi-modal, language-guided 3D affordance grounding framework based on VLM. The results show the effectiveness and better generalization capabilities of our fusion framework.
\end{itemize}

\begin{figure*}
    \centering 
    \begin{subfigure}{0.30\linewidth}
        \includegraphics[width=\linewidth]{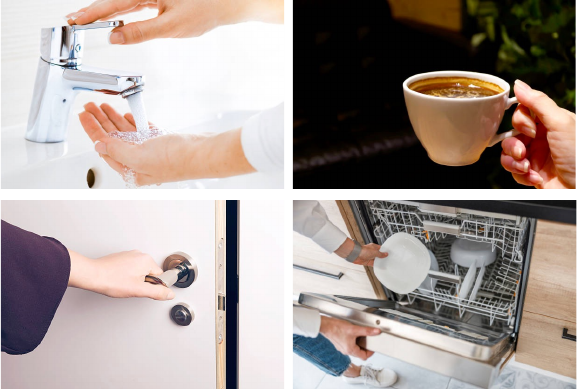}
        \caption{Example of image data.}
        \label{image_example}
    \end{subfigure}
    \quad
    \begin{subfigure}{0.355\linewidth}
        \includegraphics[width=\linewidth]{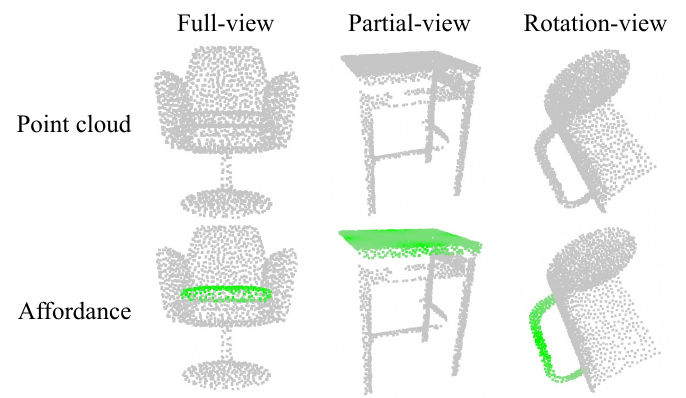}
        \caption{Example of point cloud and affordance data.}
        \label{point_example}
    \end{subfigure}
    \quad
    \begin{subfigure}{0.295\linewidth}
        \includegraphics[width=\linewidth]{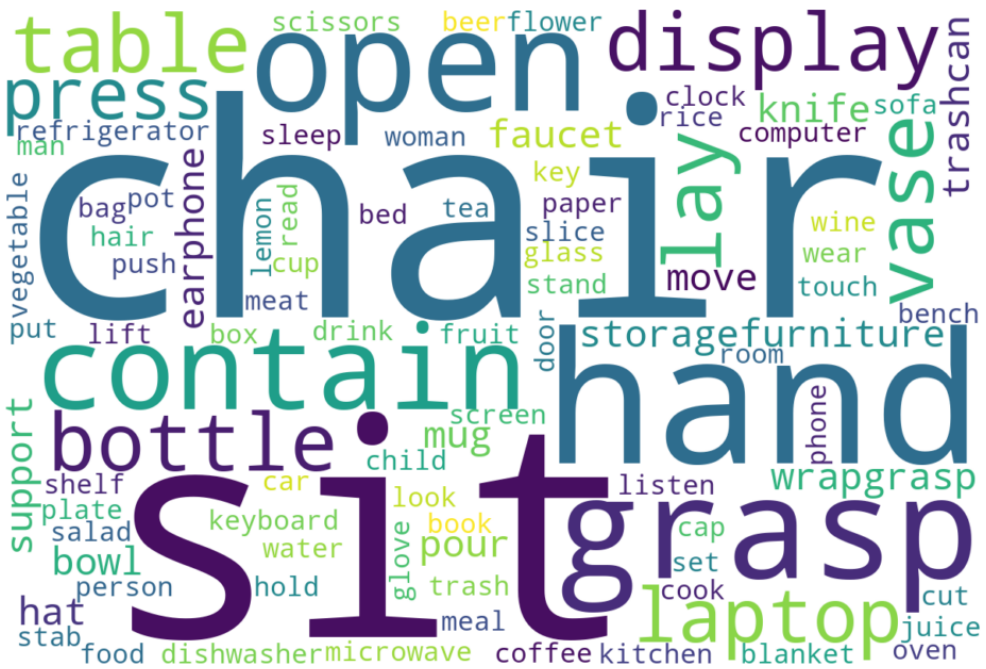}
        \caption{Distribution of instruction data.}
        \label{word_statistic}
    \end{subfigure}
    \quad
    \begin{subfigure}{0.385\linewidth}
        \includegraphics[width=\linewidth]{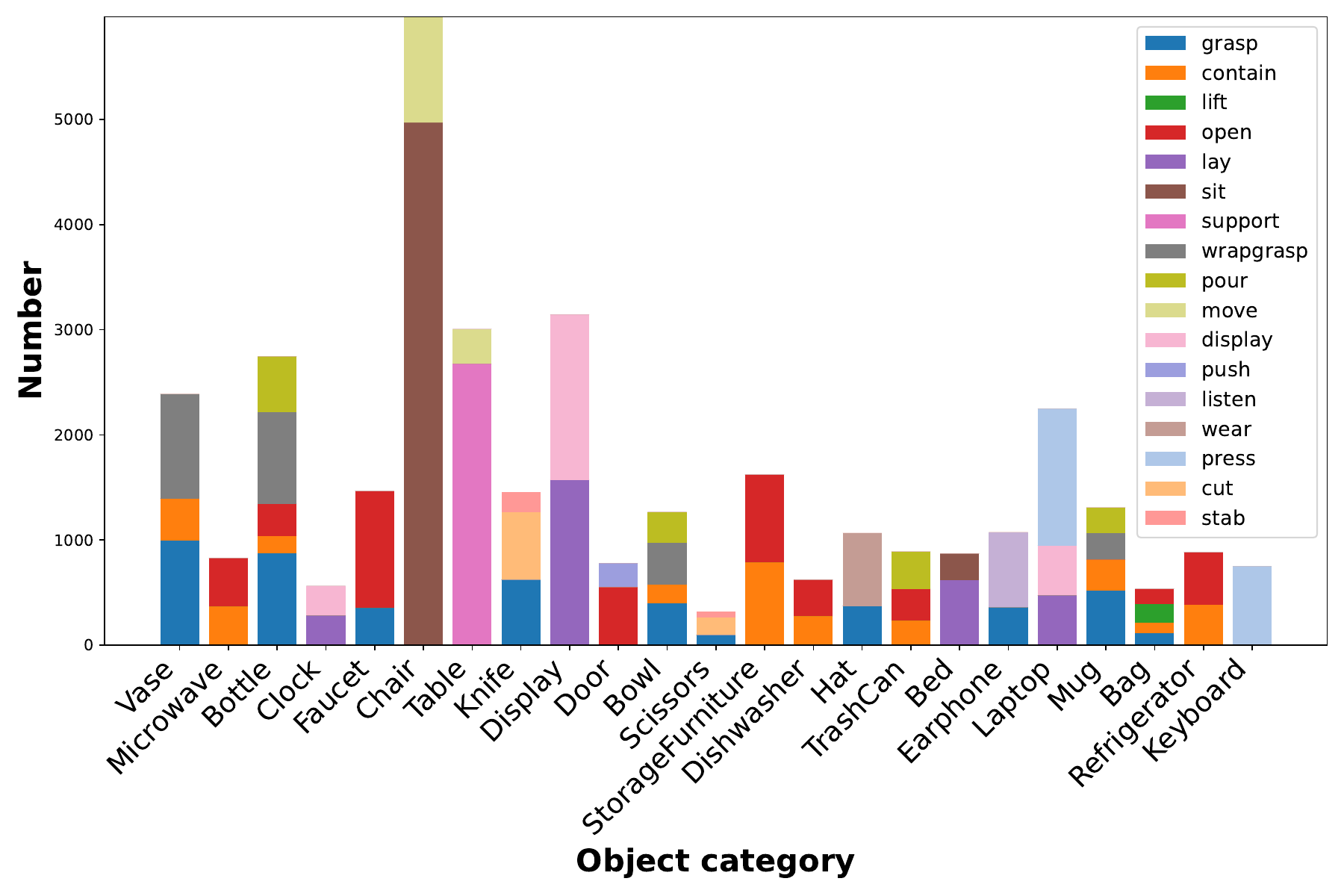}
        \caption{Distribution of image data.}
        \label{image_statistic}
    \end{subfigure}
    \quad
    \begin{subfigure}{0.385\linewidth}
        \includegraphics[width=\linewidth]{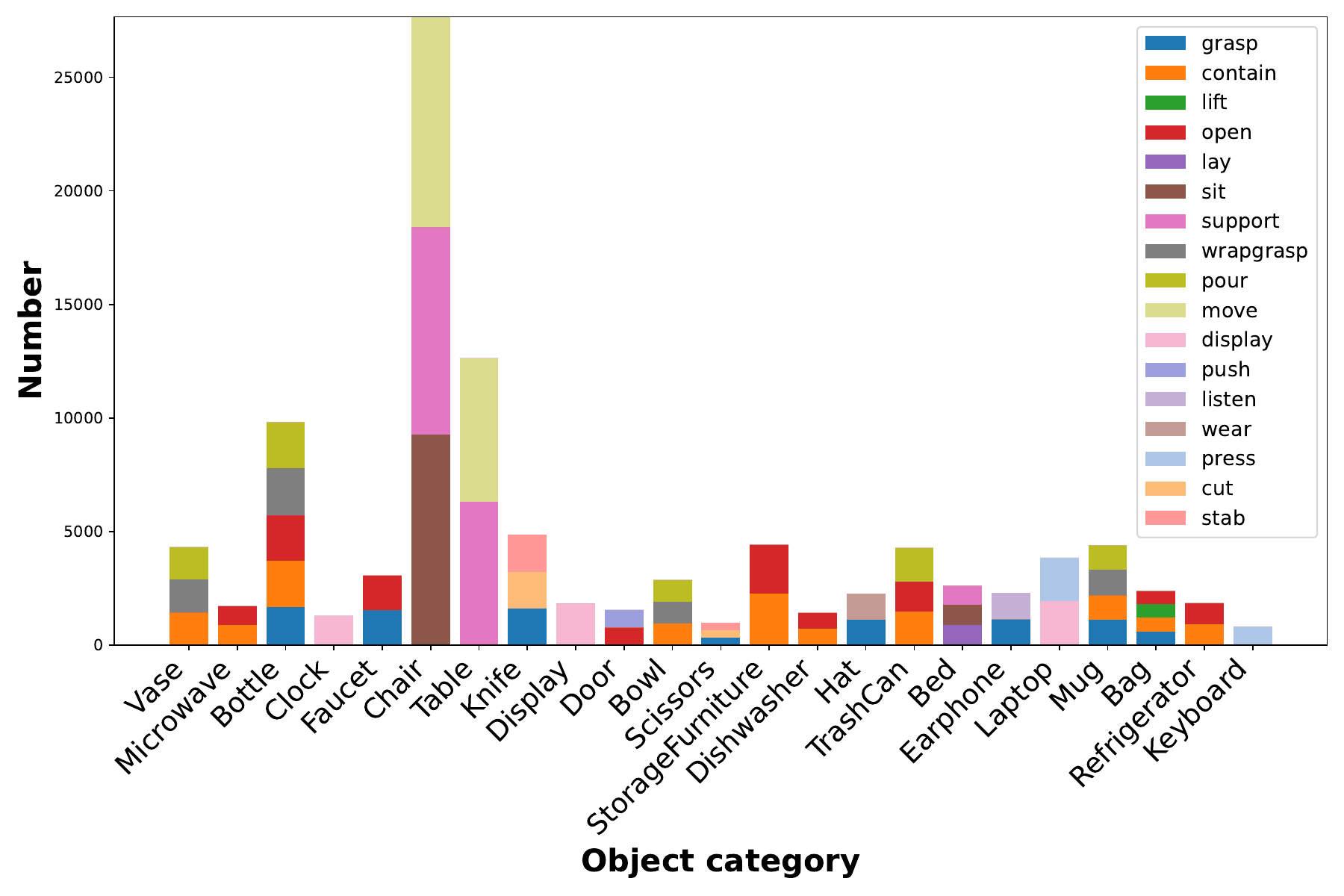}
        \caption{Distribution of point cloud data.}
        \label{point_statistic}
    \end{subfigure}
    \quad
    \begin{subfigure}{0.18\linewidth}
        \includegraphics[width=\linewidth]{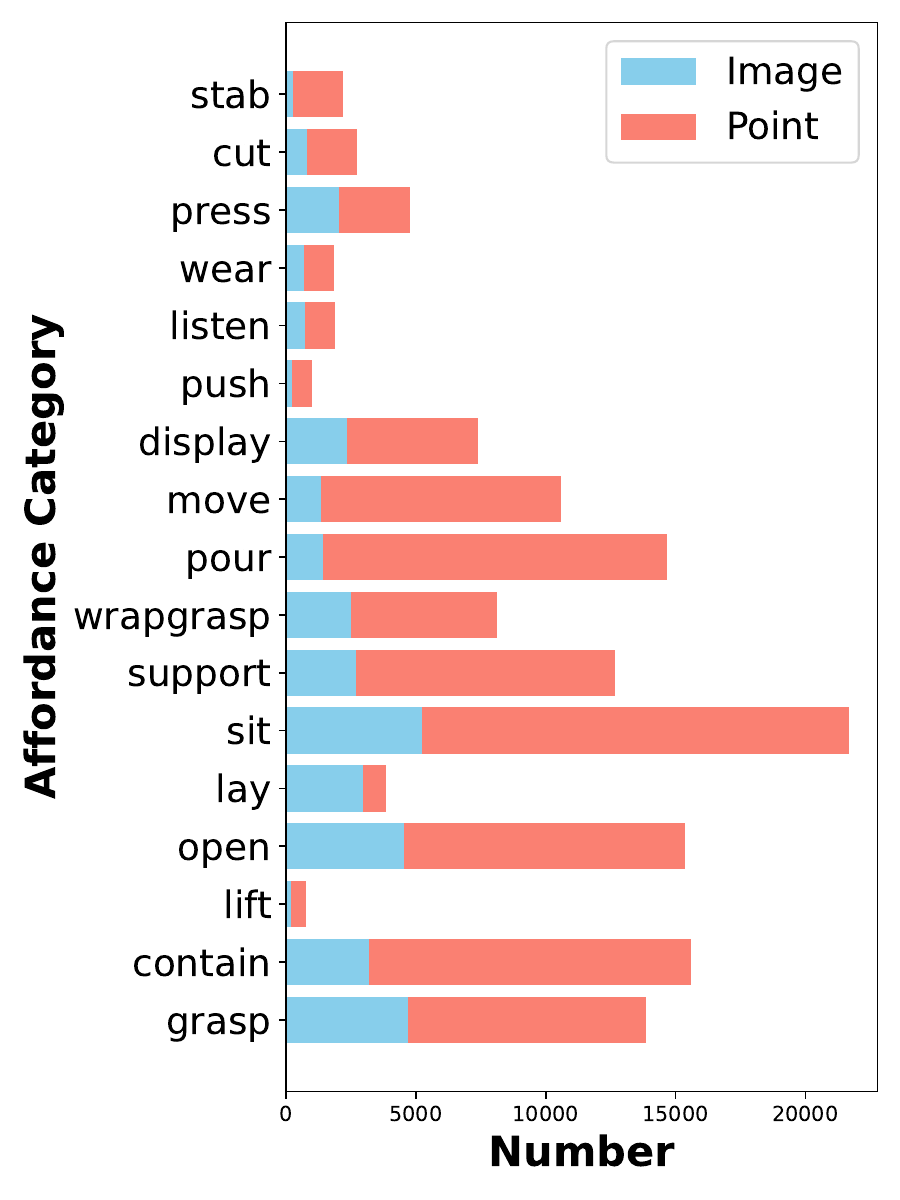}
        \caption{Distribution of affordance data.}
        \label{affordance_statistic}
    \end{subfigure}
    \caption{\textbf{Examples and statistics of the AGPIL dataset.} Figures (a) and (b) are examples of images, point clouds, and a certain affordance that we randomly selected from the dataset. Figure (c) shows a word cloud generated according to the frequency of each word appearing in the language instructions. Figures (d) and (e) respectively show the distribution of affordances corresponding to different objects in image and point cloud data. The horizontal axis represents the types of objects, and the vertical axis represents the quantity. Different colors indicate different affordances. Figure (f) illustrates the distribution of image and point cloud data corresponding to each affordance. It indicates that images and point clouds are not a one-to-one match, as a single image may correspond to multiple objects.}
    \label{statistic}
    \vspace{-0.2cm}
\end{figure*}

\section{Related Works}
\label{related_works}
\subsection{Vision-Language Models}
Due to the remarkable performance of large language models (LLMs) such as LlaMA\cite{touvron2023llama,touvron2023llama2,dubey2024llama}, Qwen\cite{bai2023qwen} and GPT\cite{radford2019language} in natural language processing tasks, many scholars have extended them to the visual domain, giving rise to a series of large vision-language models (VLMs), such as LLaVA\cite{liu2023llava, liu2023improvedllava, liu2024llavanext}, Qwen-VL\cite{bai2023qwenvl}, GPT-4o\cite{islam2024gpt}, and so on. Leveraging the prior knowledge learned by these large models, they have been applied in 3D visual grounding task. 
LLM-Grounder\cite{yang2024llm} utilizes an LLM to decompose complex natural language queries into semantic parts and employs a visual grounding tool to identify objects in a 3D scene. VLM-Grounder\cite{xu2024vlm} utilizes a VLM based on image sequences and language to achieve zero-shot 3D visual grounding. Unlike previous work on object-level grounding tasks, we define our task as utilizing multi-modal data from point clouds and images to ground 3D object affordances based on human language instructions, which is a part-level 3D grounding task.

\subsection{Affordance Grounding}
Affordance grounding is to identify the locations where interactions are possible, which is important for human-object interaction. AffordanceLLM\cite{qian2024affordancellm} uses LLaVA\cite{liu2024llavanext} for 2D visual affordance grounding based and it demonstrates the importance of 3D information in affordance grounding. However, it obtains a 2D heatmap of affordance based on the image and depth map, which is not convenient to map it to 3D space for human or robot operation. Several studies have focused on 3D affordance. For instance, 3D AffordanceNet\cite{deng20213d} predicts a 3D heatmap based on point clouds. And some methods leverage multi-modal data to predict 3D affordance, such as IAG\cite{yang2023grounding}, OpenAD\cite{nguyen2023open}, and LASO\cite{li2024laso}. However, IAG relies on the 2D detection bounding boxes of interactive objects, which limits the task’s scalability. OpenAD uses simple labels for text input, lacking support for context-rich language instructions. Moreover, LASO overlooks challenges posed by incomplete point clouds and object rotations in the real world.

 
In order to improve the generalization of grounding 3D affordance of unseen objects, various approaches can be adopted, such as contrastive learning\cite{wu2024learning}, zero-shot learning\cite{kim2024zero}, reinforcement learning\cite{nagarajan2020learning}, supervised learning based on vision-language models \cite{qian2024affordancellm} and so on. For zero-shot learning, there are issues such as incorrect HOI prompt generation and low granularity. And contrastive learning is sensitive to data augmentation strategies and is limited by similarity metric of comparing different things. Reinforcement learning typically requires a large amount of interaction data, which is time-consuming for training. In this paper, we use a vision-language model to ground 3D object affordance in full-view, partial-view and rotation-view, which can be easily extended to other new scenarios in the real world.

\begin{figure}[t]
  \centering
   \includegraphics[width=0.9\linewidth]{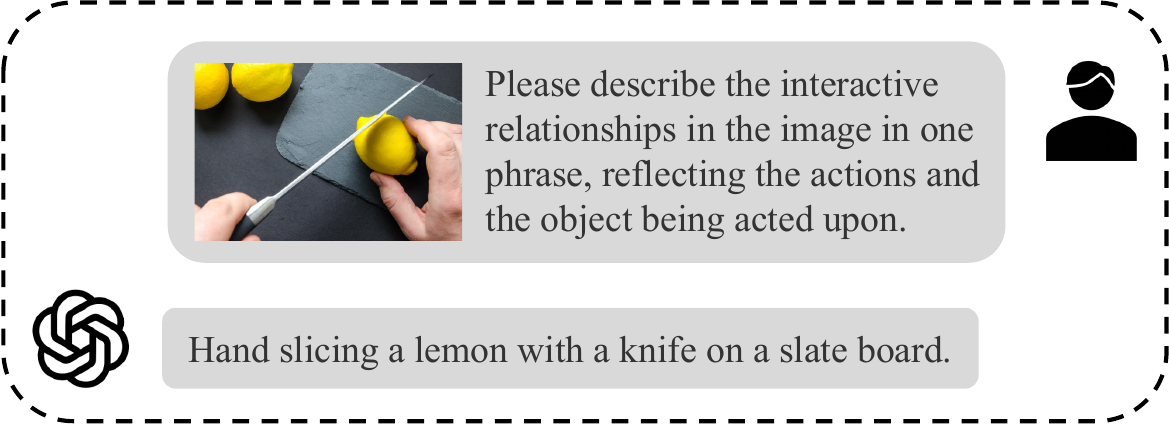}
   \caption{\textbf{Language instructions generation.} In this example, the affordance is ``cut" and the object category is ``knife".}
   \label{generation}
   \vspace{-0.36cm}
\end{figure}

\begin{figure*}[t]
  \centering
  \includegraphics[width=0.88\linewidth]{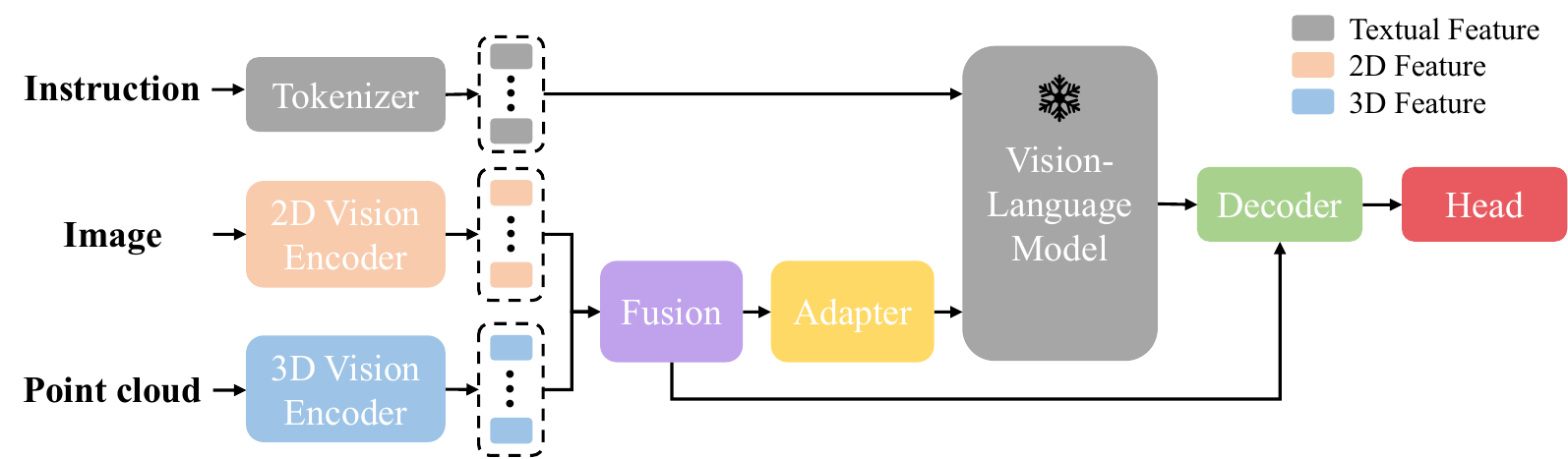}
   \caption{\textbf{Method.} The structure of the proposed LMAffordance3D model, which consists of four major components: 1) a vision encoder that processes multi-modal data, including images and point clouds, to encode and fuse the 2D and 3D features; 2) a vision-language model and its associated component (tokenizer and adapter) that takes in the instruction token, 2D and 3D vision token for fusion; 3) a decoder that uses 2D and 3D spatial features as query, instructional features as key and semantic feature as value to predict the affordance feature; 4) a head for segmenting and grounding 3D object affordance.}
   \label{model}
   \vspace{-0.36cm}
\end{figure*}

\section{Dataset}
\subsection{Collection and Generation}
We collect the 3D affordance grounding dataset (AGPIL), which contains point clouds, images and language instructions. The point clouds are mainly collected from 3D AffordanceNet\cite{deng20213d}. According to the differences in the point clouds, the dataset has three parts: \textbf{full-view}, \textbf{partial-view}, and \textbf{rotation-view}. The term ``full-view" indicates that the data include the point clouds of all the surfaces of an object. Both ``partial-view" and ``rotation-view" signify that the data include only part of the point clouds of an object's surfaces, with the distinction being whether or not there is rotation. And the rotation of the object along the x, y, and z axes is random. The images are mainly collected from AGD20K\cite{luo2024grounded} and PIAD\cite{yang2023grounding}. The image should depict the interaction between objects and humans, corresponding to the affordance of the object in the point clouds. Unlike PIAD, our dataset does not contain 2D bounding boxes of the interactive subject and object for each image. To improve efficiency, we use GPT-4o\cite{islam2024gpt} combined with an image and a carefully designed prompt to generate each language instruction. The prompt is ``Please describe the interactive relationships in the image in one phrase, reflecting the actions and the object being acted upon". To ensure accuracy, we carefully score the generated texts based on consistency, word choice accuracy, semantic completeness and conciseness, and sentence structure diversity. For texts with lower scores, we regenerate or manually refine them. The diagram which illustrates the generation of language instructions is shown in the \cref{generation}.

\subsection{Annotation}
For the affordance of objects, we use the segmented annotations from 3D AffordanceNet\cite{deng20213d}, which enables part-level prediction. And we render a 3D heatmap based on confidence scores with different colors. Each annotation is a (2048, 17) matrix, where 2048 represents the number of points in the point cloud, and 17 represents the number of affordance types. Each element in the matrix indicates the probability of a point providing a specific affordance, with a range from 0 to 1. If all elements in a column are 0.0, it indicates that the object does not have the corresponding affordance. In our task, we only use the 3D heatmap of a single type of affordance corresponding to the image and language instruction in the point cloud annotation for training. The prediction is a (2048, 1) matrix during inference. In this way, we can predict different affordances in the point clouds with different images and language instructions.

\subsection{Statistic Analysis}
As \cref{Dataset_comparion} shows, we compare the open affordance dataset with AGPIL. Our total dataset consists of 41,628 point clouds, 30,972 images, and 30,972 language instructions, encompassing 23 object categories and 17 types of affordances. Each object has an average of 2.4 functional parts. In particular, objects in the images and point clouds are not captured from the same scene and they are paired based on object categories. Our dataset has two settings: \textbf{Seen} and \textbf{Unseen}. In the seen dataset, the categories of objects and affordances in the training and test sets are consistent, while in the unseen dataset, they are inconsistent. More examples and statistics about the dataset are illustrated in \cref{statistic}.

\section{Method}
\label{method}
\subsection{Overview}
In this section, we propose LMAffordance3D, a framework that learns from language instructions, visual observations and interactions for grounding 3D object affordance. As illustrated in \cref{model}, LMAffordance3D is composed of four major components: 1) a vision encoder that processes multi-modal data, including images and point clouds, to encode and fuse the 2D and 3D features, where an RGB image contains information about the colors, scenes, and interactions between objects, while point clouds contain different information about the shape, size, and geometry of objects. 2) a vision-language model and its associated components (tokenizer and adapter) that take in the instruction token, 2D and 3D vision token for fusion; 3) a decoder that uses 2D and 3D spatial features as query, instructional features as key and semantic feature as value to predict the affordance feature; 4) a head for segmenting and grounding 3D object affordance. We will introduce the components of LMAffordance3D in \cref{LMAffordance3D}. And we present the loss function in \cref{loss_function}.

\begin{table*}[ht]
    \centering
    \resizebox{\textwidth}{!}{%
    \begin{tabular}{c|r|ccccc|ccccc|ccccc}
        \toprule
        \multirow{2}{*}{\textbf{Setting}} & \multirow{2}{*}{\textbf{Metrics}} &
        \multicolumn{5}{c|}{\textbf{Full-view}} & 
        \multicolumn{5}{c|}{\textbf{Partial-view}} & 
        \multicolumn{5}{c}{\textbf{Rotation-view}} \\
        \cmidrule(r){3-7} \cmidrule(r){8-12} \cmidrule(r){13-17}
        & & \textbf{3D A.N.} & \textbf{IAG} & \textbf{O.AD} & \textbf{P.R.} & \textbf{Ours} & 
        \textbf{3D A.N.} & \textbf{IAG} & \textbf{O.AD} & \textbf{P.R.} & \textbf{Ours} & 
        \textbf{3D A.N.} & \textbf{IAG} & \textbf{O.AD} & \textbf{P.R.} & \textbf{Ours} \\
        \midrule
        \multirow{4}{*}{\textbf{Seen}} 
        & AUC $\uparrow$ & 0.8067 & 0.8485 & 0.8584 & 0.8766 & 0.8895 & 0.7609 & 0.8085 & 0.8145 & 0.8212 & 0.8478 & 0.5954 & 0.6791 & 0.7325 & 0.7564 & 0.7823 \\
        & IOU $\uparrow$  & 0.1341 & 0.2051 & 0.1993 & 0.2002 & 0.2123 & 0.1077 & 0.1454 & 0.1562 & 0.1613 & 0.1755 & 0.0452 & 0.0804 & 0.0911 & 0.1012 & 0.1161 \\
        & SIM $\uparrow$  & 0.4833 & 0.5450 & 0.5866 & 0.5953 & 0.6102 & 0.4785 & 0.5362 & 0.5652 & 0.5796 & 0.5928 & 0.3892 & 0.4359 & 0.4783 & 0.4984 & 0.5191 \\
        & MAE $\downarrow$  & 0.1141 & 0.0980 & 0.0944 & 0.0905 & 0.0816 & 0.1262 & 0.1042 & 0.1006 & 0.0987 & 0.0921 & 0.1648 & 0.1345 & 0.1236 & 0.1302 & 0.1182 \\
        \midrule
        \multirow{4}{*}{\textbf{Unseen}} 
        & AUC $\uparrow$  & 0.6652 & 0.7184 & 0.7375 & 0.7545 & 0.7741 & 0.6419 & 0.6718 & 0.7014 & 0.7211 & 0.7602 & 0.5821 & 0.5994 & 0.5989 & 0.6088 & 0.6303 \\
        & IOU $\uparrow$  & 0.0522 & 0.0795 & 0.0781 & 0.0813 & 0.0903 & 0.0425 & 0.0612 & 0.0671 & 0.0682 & 0.0724 & 0.0277 & 0.0311 & 0.0356 & 0.0379 & 0.0415 \\
        & SIM $\uparrow$  & 0.3012 & 0.3520 & 0.3704 & 0.3849 & 0.4089 & 0.3347 & 0.3568 & 0.3766 & 0.3894 & 0.4144 & 0.3552 & 0.3623 & 0.3622 & 0.3631 & 0.3842 \\
        & MAE $\downarrow$  & 0.1556 & 0.1270 & 0.1193 & 0.1178 & 0.0945 & 0.1601 & 0.1518 & 0.1428 & 0.1262 & 0.1183 & 0.1744 & 0.1709 & 0.1506 & 0.1503 & 0.1398 \\
        \bottomrule
    \end{tabular}
    }
    \caption{\textbf{Benchmark.} We present the overall results of all comparative methods on different views and settings. ``3D A.N.'' refers to ``3D AffordanceNet'', ``O.AD'' refers to ``OpenAD'', and ``P.R.'' refers to ``PointRefer''}
    \label{results}
    \vspace{-0.36cm}
\end{table*}

\subsection{LMAffordance3D}
\label{LMAffordance3D}
\textbf{Vision encoder.} In the vision-language community\cite{li2022blip, li2023blip, liu2023llava, liu2023improvedllava, liu2024llavanext, zhu2023minigpt}, the most common method for vision-language alignment is to use a pre-trained CLIP\cite{radford2021learning} model and project image features into the semantic space. However, we need to process images $I_I \in \mathbb{R}^{B\times 3\times H \times W}$ and point clouds $I_P \in \mathbb{R}^{B\times 3 \times 2048}$ data for fusion in this task. And the large flops and parameter size of the CLIP models pose challenges for deployment in robotics. So we design a multi-modal vision encoder, in which the 2D vision encoder uses ResNet18\cite{he2016deep} to extract 2D features $F_{2D} \in \mathbb{R}^{B\times C_I \times H \times W}$, and the 3D vision encoder uses PointNet++\cite{qi2017pointnet++} to extract 3D features $F_{3D} \in \mathbb{R}^{B\times C_P \times N_P}$. Then fusion module uses MLP and self-attention to obtain multi-modal spatial features $F_S \in \mathbb{R}^{B\times N_S \times C_S}$. 

\textbf{Vision-language model.} Vision-language model as the brain of agent is used for comprehending natural language instructions, integrating semantic features and spatial features to obtain affordance features. Specifically, we choose LLaVA-7B as the backbone networks for the vision-language model. Given the action instruction, we apply the tokenizer to convert the instruction into textual feature $F_T \in \mathbb{R}^{B\times N_L \times C_L}$. We use an adapter, which consists of two linear layers and an activation layer. It takes spatial features $F_S \in \mathbb{R}^{B\times N_S \times C_S}$ as input and maps them to a semantic space while making an increase in the channel dimension to obtain projected features $F_{SP} \in \mathbb{R}^{B\times N_S \times C_L}$. These projected features $F_{SP}$ are then concatenated with textual features $F_{T}$ as the input of vision-language model. And the output of vision-language model is its hidden state.

\textbf{Decoder.} In order to fuse all kinds of feature to get the affordance features, we design a decoder based on cross-attention. We split the output of vision-language model into instructional features and semantic features. Then we use spatial features as the query, instructional features as the key, and semantic features as the value, and obtain the affordance feature $F_A \in \mathbb{R}^{B\times N_A \times C_A}$ with the decoder.

\textbf{Head.} Here we upsample the affordance feature $F_A$ for propagation and use a head composed of two linear layers, a activation layer, a batch normalization layer and a sigmoid layer to obtain 3D object affordance $O \in \mathbb{R}^{B\times 2048 \times 1}$.

\subsection{Loss Function}
\label{loss_function}
We predict the probability that each point cloud belongs to the corresponding affordance, ranging from 0 to 1. In order to supervise the point-wise 3D heatmap on point clouds, we use a weighted sum of focal loss\cite{ross2017focal} and dice loss\cite{milletari2016v} as the final loss function:

\begin{equation}
  Loss = \omega_{f} L_{f} + \omega_{d} L_{d}
  \label{loss}
\end{equation}

where $\omega_{f}$ and $\omega_{d}$ represent the weight of the focal loss and dice loss respectively.

\section{Experiments}
\begin{table*}[ht]
    \centering
    \resizebox{\textwidth}{!}{%
    \begin{tabular}{c|r|ccccccccccccccccc}
        \toprule
         \textbf{F} & \textbf{Metrics} & \textbf{grasp} & \textbf{press} & \textbf{stab} & \textbf{open} & \textbf{lay} & \textbf{sit} & \textbf{cut} & \textbf{cont.} & \textbf{disp.} & \textbf{wrap.} & \textbf{supp.} & \textbf{push} & \textbf{listen} & \textbf{wear} & \textbf{move} & \textbf{lift} & \textbf{pour} \\
        \midrule
        \multirow{4}{*}{\rotatebox{90}{Seen}} 
        & AUC$\uparrow$ & 0.859 & 0.898 & 0.997 & 0.932 & 0.935 & 0.957 & 0.945 & 0.871 & 0.907 & 0.689 & 0.846 & 0.801 & 0.910 & 0.733 & 0.776 & 0.973 & 0.945 \\
        & aIOU$\uparrow$ & 0.203 & 0.136 & 0.354 & 0.239 & 0.294 & 0.358 & 0.156 & 0.199 & 0.293 & 0.060 & 0.111 & 0.029 & 0.160 & 0.052 & 0.098 & 0.407 & 0.232 \\
        & SIM$\uparrow$ & 0.650 & 0.401 & 0.651 & 0.398 & 0.693 & 0.719 & 0.681 & 0.567 & 0.655 & 0.645 & 0.716 & 0.570 & 0.683 & 0.564 & 0.612 & 0.497 & 0.597 \\
        & MAE$\downarrow$ & 0.098 & 0.059 & 0.017 & 0.050 & 0.086 & 0.065 & 0.056 & 0.089 & 0.095 & 0.128 & 0.097 & 0.082 & 0.087 & 0.138 & 0.135 & 0.020 & 0.075 \\
        \midrule
        \multirow{4}{*}{\rotatebox{90}{Unseen}} 
        & AUC$\uparrow$ & 0.363 & 0.854 & 0.932 & 0.894 & 0.788 & 0.719 & 0.921 & 0.866 & 0.681 & 0.556 & \textbf{-} & \textbf{-} & \textbf{-} & \textbf{-} & \textbf{-} & \textbf{-} & \textbf{-} \\
        & aIOU$\uparrow$ & 0.008 & 0.038 & 0.053 & 0.103 & 0.122 & 0.109 & 0.104 & 0.150 & 0.047 & 0.025 & \textbf{-} & \textbf{-} & \textbf{-} & \textbf{-} & \textbf{-} & \textbf{-} & \textbf{-} \\
        & SIM$\uparrow$ & 0.319 & 0.304 & 0.356 & 0.215 & 0.427 & 0.376 & 0.533 & 0.522 & 0.277 & 0.562 & \textbf{-} & \textbf{-} & \textbf{-} & \textbf{-} & \textbf{-} & \textbf{-} & \textbf{-} \\
        & MAE$\downarrow$ & 0.205 & 0.065 & 0.053 & 0.058 & 0.127 & 0.110 & 0.059 & 0.098 & 0.111 & 0.134 & \textbf{-} & \textbf{-} & \textbf{-} & \textbf{-} & \textbf{-} & \textbf{-} & \textbf{-} \\
        \bottomrule
         \textbf{P} & \textbf{Metrics} & \textbf{grasp} & \textbf{press} & \textbf{stab} & \textbf{open} & \textbf{lay} & \textbf{sit} & \textbf{cut} & \textbf{cont.} & \textbf{disp.} & \textbf{wrap.} & \textbf{supp.} & \textbf{push} & \textbf{listen} & \textbf{wear} & \textbf{move} & \textbf{lift} & \textbf{pour} \\
        \midrule
        \multirow{4}{*}{\rotatebox{90}{Seen}} 
        & AUC$\uparrow$ & 0.818 & 0.836 & 0.987 & 0.863 & 0.791 & 0.952 & 0.909 & 0.806 & 0.857 & 0.647 & 0.832 & 0.781 & 0.858 & 0.709 & 0.811 & 0.963 & 0.907 \\
        & aIOU$\uparrow$ & 0.151 & 0.127 & 0.377 & 0.212 & 0.134 & 0.313 & 0.155 & 0.128 & 0.264 & 0.041 & 0.086 & 0.029 & 0.103 & 0.036 & 0.095 & 0.393 & 0.214 \\
        & SIM$\uparrow$ & 0.608 & 0.411 & 0.522 & 0.404 & 0.465 & 0.702 & 0.751 & 0.498 & 0.602 & 0.744 & 0.731 & 0.549 & 0.510 & 0.657 & 0.622 & 0.500 & 0.637 \\
        & MAE$\downarrow$ & 0.115 & 0.078 & 0.035 & 0.074 & 0.101 & 0.058 & 0.065 & 0.103 & 0.117 & 0.137 & 0.102 & 0.110 & 0.099 & 0.135 & 0.119 & 0.063 & 0.089 \\
        \midrule
        \multirow{4}{*}{\rotatebox{90}{Unseen}} 
        & AUC$\uparrow$ & 0.358 & 0.748 & 0.939 & 0.880 & 0.801 & 0.742 & 0.879 & 0.811 & 0.577 & 0.604 & \textbf{-} & \textbf{-} & \textbf{-} & \textbf{-} & \textbf{-} & \textbf{-} & \textbf{-} \\
        & aIOU$\uparrow$ & 0.004 & 0.028 & 0.057 & 0.100 & 0.141 & 0.093 & 0.047 & 0.092 & 0.043 & 0.043 & \textbf{-} & \textbf{-} & \textbf{-} & \textbf{-} & \textbf{-} & \textbf{-} & \textbf{-} \\
        & SIM$\uparrow$ & 0.358 & 0.254 & 0.429 & 0.271 & 0.451 & 0.362 & 0.450 & 0.514 & 0.232 & 0.621 & \textbf{-} & \textbf{-} & \textbf{-} & \textbf{-} & \textbf{-} & \textbf{-} & \textbf{-} \\
        & MAE$\downarrow$ & 0.092 & 0.106 & 0.088 & 0.093 & 0.126 & 0.120 & 0.092 & 0.116 & 0.188 & 0.153 & \textbf{-} & \textbf{-} & \textbf{-} & \textbf{-} & \textbf{-} & \textbf{-} & \textbf{-} \\
        \bottomrule
        \textbf{R} & \textbf{Metrics} & \textbf{grasp} & \textbf{press} & \textbf{stab} & \textbf{open} & \textbf{lay} & \textbf{sit} & \textbf{cut} & \textbf{cont.} & \textbf{disp.} & \textbf{wrap.} & \textbf{supp.} & \textbf{push} & \textbf{listen} & \textbf{wear} & \textbf{move} & \textbf{lift} & \textbf{pour} \\
        \midrule
        \multirow{4}{*}{\rotatebox{90}{Seen}} 
        & AUC$\uparrow$ & 0.758 & 0.726 & 0.837 & 0.785 & 0.820 & 0.868 & 0.835 & 0.683 & 0.778 & 0.608 & 0.752 & 0.861 & 0.776 & 0.586 & 0.765 & 0.777 & 0.829 \\
        & aIOU$\uparrow$ & 0.125 & 0.042 & 0.178 & 0.108 & 0.106 & 0.190 & 0.102 & 0.041 & 0.190 & 0.032 & 0.069 & 0.022 & 0.097 & 0.018 & 0.055 & 0.246 & 0.145 \\
        & SIM$\uparrow$ & 0.594 & 0.259 & 0.562 & 0.294 & 0.461 & 0.537 & 0.650 & 0.432 & 0.567 & 0.706 & 0.659 & 0.594 & 0.543 & 0.648 & 0.599 & 0.466 & 0.541 \\
        & MAE$\downarrow$ & 0.116 & 0.117 & 0.065 & 0.106 & 0.161 & 0.098 & 0.103 & 0.135 & 0.165 & 0.138 & 0.136 & 0.081 & 0.103 & 0.128 & 0.136 & 0.057 & 0.119 \\
        \midrule
        \multirow{4}{*}{\rotatebox{90}{Unseen}} 
        & AUC$\uparrow$ & 0.788 & 0.643 & 0.992 & 0.565 & 0.760 & 0.602 & 0.925 & 0.576 & 0.533 & 0.583 & \textbf{-} & \textbf{-} & \textbf{-} & \textbf{-} & \textbf{-} & \textbf{-} & \textbf{-} \\
        & aIOU$\uparrow$ & 0.029 & 0.024 & 0.422 & 0.006 & 0.110 & 0.045 & 0.123 & 0.027 & 0.055 & 0.031 & \textbf{-} & \textbf{-} & \textbf{-} & \textbf{-} & \textbf{-} & \textbf{-} & \textbf{-} \\
        & SIM$\uparrow$ & 0.589 & 0.208 & 0.587 & 0.168 & 0.431 & 0.337 & 0.628 & 0.485 & 0.249 & 0.656 & \textbf{-} & \textbf{-} & \textbf{-} & \textbf{-} & \textbf{-} & \textbf{-} & \textbf{-} \\
        & MAE$\downarrow$ & 0.101 & 0.144 & 0.028 & 0.082 & 0.173 & 0.175 & 0.079 & 0.147 & 0.233 & 0.152 & \textbf{-} & \textbf{-} & \textbf{-} & \textbf{-} & \textbf{-} & \textbf{-} & \textbf{-} \\
        \bottomrule
    \end{tabular}
    }
    \caption{\textbf{Evaluation metrics for different affordance with LMAffordance3D.} We compare on seen and unseen setting in different views, where F represents full-view, P represents partial-view, and R represents rotation-view. ``cont." refers to ``contain", ``disp." refers to ``display", ``wrap." refers to ``wrapgrasp", and ``supp." refers to ``support". ``\textbf{-}" indicates that there is no corresponding affordance category in the test set, hence no metrics.}
    \label{affordance_analysis}
    \vspace{-0.36cm}
\end{table*}

\subsection{Benchmark}
\textbf{Metrics.} In accordance with PIAD\cite{yang2023grounding}, we use four metrics to build the AGPIL benchmark: AUC\cite{lobo2008auc}, aIoU\cite{rahman2016optimizing}, SIM\cite{swain1991color}, and MAE\cite{willmott2005advantages}. The AUC computes the area under the ROC curve to assess the significance map on a 3D point cloud heatmap. The aIOU is the average intersection over union, which is used to evaluate the proximity between the affordance prediction and the ground truth. SIM measures the similarity between the predicted map and the ground truth map. MAE is the mean absolute error between the predicted probabilities and the ground truth. For AUC, aIoU, and SIM, higher is better, while for MAE, lower is better.

\textbf{Baselines.} Since there are no prior works on grounding 3D object affordance based on language, images, and point clouds, we choose 3D AffordanceNet\cite{qi2017pointnet++}, IAG\cite{yang2023grounding}, OpenAD\cite{nguyen2023open}, and PointRefer\cite{li2024laso} as baselines. We evaluate them in both seen and unseen settings, as well as under different perspectives including full-view, partial-view, and rotation-view. This allows us to comprehensively understand the model's generalization ability and performance across different object categories and viewpoints.

\textbf{Training and testing setup.} Our model is implemented using PyTorch and all training and testing experiments are conducted on a single V100 GPU. We train the model with the AdamW optimizer\cite{loshchilov2017decoupled} and a cosine learning rate scheduler\cite{loshchilov2016sgdr}. The initial learning rate is 1e-4 and the weight decay is set to 0.06. We train the models with the first 2000 iteration steps for warm-up\cite{he2016deep}. The number of training epochs is set to 20 and the training batch size is set to 6. The image 2D vision encoder ResNet18 uses parameters pre-trained on ImageNet\cite{deng2009imagenet}, while the point cloud 3D vision encoder is trained from scratch. The vision-language model uses llava-v1.6-vicuna-7b, and its parameters are frozen. During training, automatic mixed precision (AMP) is introduced to accelerate training and reduce memory usage. Furthermore, given that images and point clouds do not need to be strictly one-to-one paired, we pair images and point clouds online during training. In one training step, an image can be paired with two point clouds, effectively augmenting the training samples. During training, the loss of one image and all paired point clouds is accumulated, and gradients are calculated based on the accumulated loss. During inference, an image is paired with only one point cloud to predict the result of affordance.

\begin{figure*}[ht]
    \centering
    \includegraphics[width=0.95\linewidth]{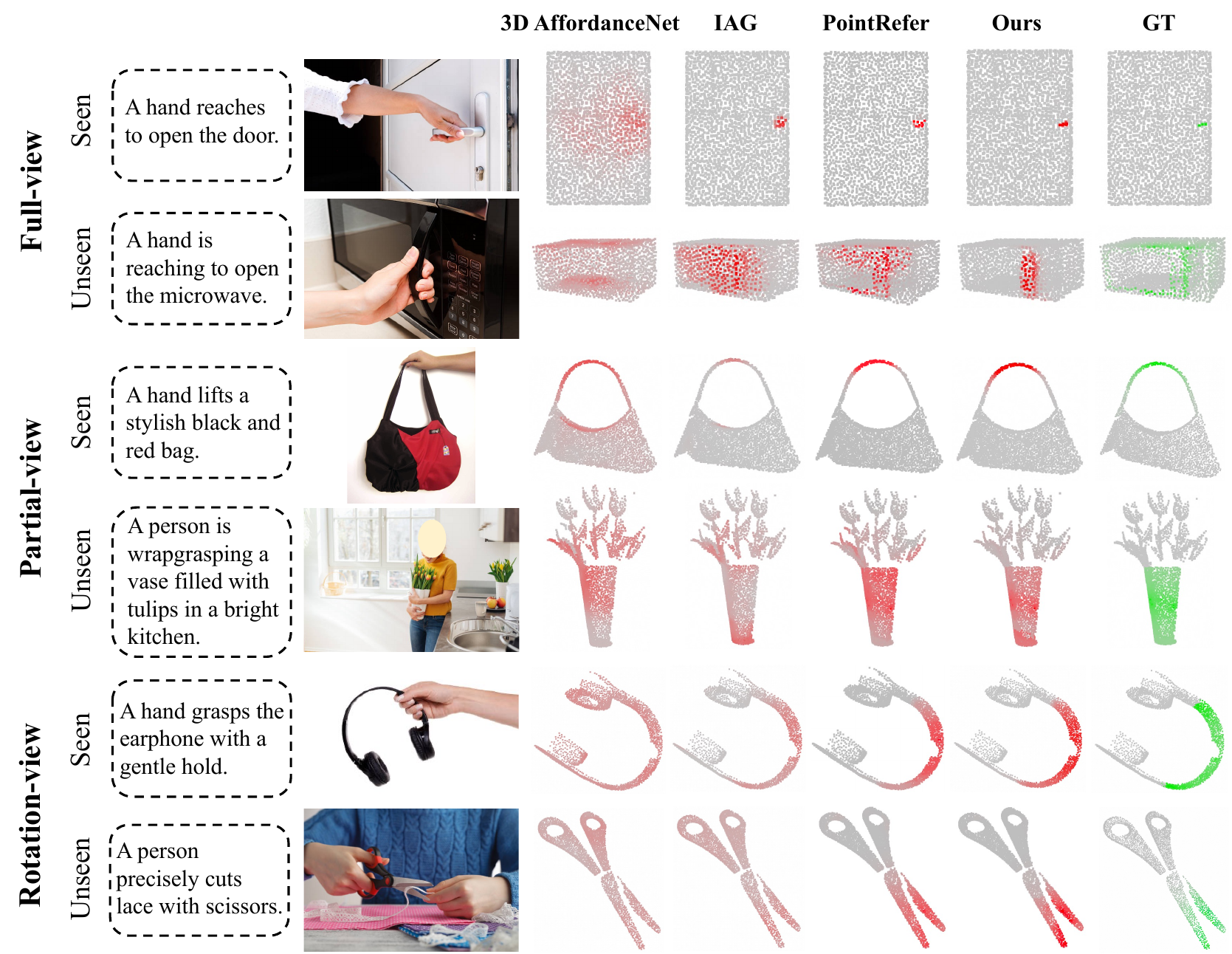}
    \caption{\textbf{Visualization.} We select several examples from the test set under different views and experimental settings, showcasing the model’s inputs and outputs, and comparing them with the ground truth (GT).}
    \label{visualization}
    \vspace{-0.36cm}
\end{figure*}

\subsection{Results and Analysis} 
\textbf{Quantitative results.}
We build a benchmark based on the AGPIL dataset, as shown in the \cref{results}. It can be observed that our method outperforms the compared baselines, and there is a significant performance improvement in the unseen setting compared to the seen setting. This improvement can be attributed to the extensive world knowledge and prior information in the large language model. In the same experimental setup, the metrics for the full view are greater than those for the partial view, which in turn are greater than those for the rotation view. In the \cref{affordance_analysis}, we present the evaluation metrics for different affordances using LMAffordance3D in detail. In the seen setting, both the training and testing datasets include all affordance categories. However, in the unseen setting, the training dataset is limited to seven categories: support, push, listen, wear, move, lift, and pour. The test dataset, on the other hand, consists of ten categories: grasp, press, stab, open, lay, sit, cut, contain, display, and wrapgrasp. The results indicate that our model demonstrates good generalization performance.

\textbf{Qualitative analysis.}
As shown in the \cref{visualization}, we compare the results of affordance prediction using different methods across various views and settings. When in full view, the model is able to observe the object from various angles, maximizing its grasp of the geometric and morphological information of the object. This setting can provide the model with the most complete information, allowing for comprehensive affordance analysis. In partial view, the model can only observe a partial view of the object, simulating real-world situations where full information about the object is inaccessible due to occlusion or viewpoint limitations. This experiment tests the model's robustness when information is incomplete. In rotation view, the object is rotated, and only partial views of the object are observed. It evaluates the model's robustness in handling object pose rotations or changes in observer perspective. The figure clearly shows that our method achieves significantly better results, confirming the validity of our setting and demonstrating the superiority of our method.

\subsection{Ablation Study}
We conduct ablation experiments on the AGPIL dataset to evaluate the influence of interaction images and language instructions on the model. The results are shown in \cref{ablation}. 

To assess whether the visual cues from RGB images affect affordance accuracy, we can find that the performance without image input decreases in both seen and unseen settings, with a significant drop in the unseen setting. This demonstrates the necessity of interactive image input for this task.

To explore the impact of different language instructions on the experiment, we select four different settings:
\begin{itemize}[itemsep=0.66pt, topsep=2pt]
\item \textbf{Full}: We use the content generated by GPT-4o as the instruction.
\item \textbf{Action \& Object}: We use an action and an object name, that is, a verb and a noun as an instruction, such as ``grasp bottle".
\item \textbf{Action}: The instruction includes only an action, for example, ``grasp".
\item \textbf{None}: An empty instruction.
\end{itemize}
We can observe that full instructions yield a higher performance compared to other simpler prompts. This demonstrates that full instructions contain more knowledge and information, which are more beneficial for the pre-trained large language model to understand the intent.

\begin{table*}[ht]
    \centering
    \renewcommand{\arraystretch}{0.66} 
    \begin{tabular}{c|c|cc|cccc}
        \toprule
        \textbf{View} & \textbf{Setting} & \textbf{Image} & \textbf{Instruction} & \textbf{AUC$\uparrow$} & \textbf{aIOU$\uparrow$} & \textbf{SIM$\uparrow$} & \textbf{MAE$\downarrow$} \\
        \midrule
        \multirow{10}{*}{\textbf{Full-view}} 
        & \multirow{5}{*}{Seen} 
        & \cmark & Full             & \textbf{0.8895} & \textbf{0.2123} & \textbf{0.6102} & \textbf{0.0816} \\
        &  & \xmark & Full          & 0.8827 & 0.2037 & 0.6045 & 0.0881 \\
        &  & \cmark & Action \& Object & 0.8829 & 0.2057 & 0.6058 & 0.0829 \\
        &  & \cmark & Action        & 0.8813 & 0.2046 & 0.6034 & 0.0842 \\
        &  & \cmark & None          & 0.8263 & 0.1705 & 0.5394 & 0.1018 \\
        \cmidrule{2-8}
        & \multirow{5}{*}{Unseen} 
        & \cmark & Full             & \textbf{0.7741} & \textbf{0.0903} & \textbf{0.4089} & \textbf{0.0945} \\
        &  & \xmark & Full          & 0.7604 & 0.0856 & 0.3990 & 0.1213 \\
        &  & \cmark & Action \& Object & 0.7712 & 0.0783 & 0.4008 & 0.0968 \\ 
        &  & \cmark & Action        & 0.7667 & 0.0774 & 0.3980 & 0.0997 \\
        &  & \cmark & None          & 0.6741 & 0.0371 & 0.3036 & 0.1282 \\
        \midrule
        \multirow{10}{*}{\textbf{Partial-view}} 
        & \multirow{5}{*}{Seen} 
        & \cmark & Full             & \textbf{0.8478} & \textbf{0.1755} & \textbf{0.5928} & \textbf{0.0921} \\ 
        &  & \xmark & Full          & 0.8383 & 0.1642 & 0.5854 & 0.0972 \\
        &  & \cmark & Action \& Object & 0.8457 & 0.1721 & 0.5882 & 0.0924 \\
        &  & \cmark & Action        & 0.8425 & 0.1706 & 0.5866 & 0.0937 \\
        &  & \cmark & None          & 0.8281 & 0.1512 & 0.5702 & 0.0996 \\
        \cmidrule{2-8}
        & \multirow{5}{*}{Unseen} 
        & \cmark & Full             & \textbf{0.7602} & \textbf{0.0724} & \textbf{0.4144} & \textbf{0.1183} \\ 
        &  & \xmark & Full          & 0.7426 & 0.0701 & 0.4022 & 0.1311 \\
        &  & \cmark & Action \& Object & 0.7600 & 0.0692 & 0.4096 & 0.1216 \\ 
        &  & \cmark & Action        & 0.7587 & 0.0598 & 0.4070 & 0.1198 \\ 
        &  & \cmark & None          & 0.6513 & 0.0475 & 0.3412 & 0.1566 \\
        \midrule
        \multirow{10}{*}{\textbf{Rotation-view}} 
        & \multirow{5}{*}{Seen} 
        & \cmark & Full             & \textbf{0.7823} & 0.1161 & \textbf{0.5191} & \textbf{0.1182} \\
        &  & \xmark & Full          & 0.7732 & 0.1063 & 0.5067 & 0.1214 \\
        &  & \cmark & Action \& Object & 0.7742 & \textbf{0.1163} & 0.5188 & 0.1238 \\
        &  & \cmark & Action        & 0.7731 & 0.1159 & 0.5155 & 0.1189 \\
        &  & \cmark & None          & 0.7052 & 0.085 & 0.4657 & 0.1394 \\
        \cmidrule{2-8}
        & \multirow{5}{*}{Unseen} 
        & \cmark & Full             & \textbf{0.6303} & \textbf{0.0415} & \textbf{0.3842} & 0.1398 \\
        &  & \xmark & Full          & 0.6126 & 0.0401 & 0.3709 & 0.1512 \\
        &  & \cmark & Action \& Object & 0.6283 & 0.0413 & 0.3794 & \textbf{0.1390} \\
        &  & \cmark & Action        & 0.6277 & 0.0412 & 0.3782 & 0.1403 \\
        &  & \cmark & None          & 0.5899 & 0.0369 & 0.3702 & 0.1489 \\
        \bottomrule
    \end{tabular}
    \caption{\textbf{Ablation Study.} We investigate the impact of interaction images and language instructions on model performance relative to the baseline, with the best results highlighted in \textbf{bold}.}
    \label{ablation}
    \vspace{-0.25cm}
\end{table*}

\subsection{Generalization to Real World}

\begin{figure}[t]
  \centering
   \includegraphics[width=0.953\linewidth]{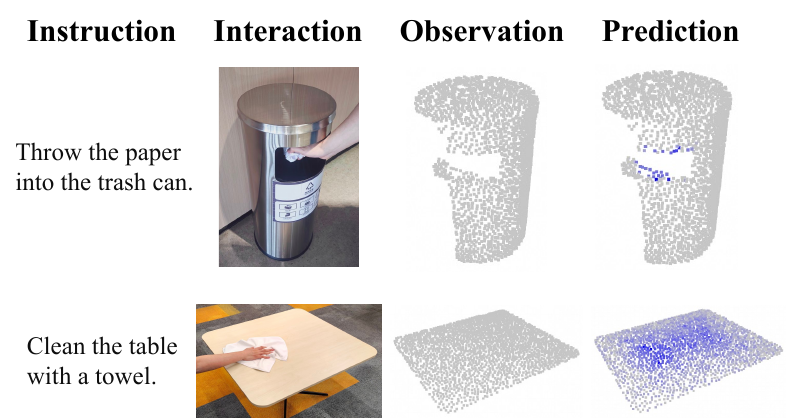}
   \caption{\textbf{Real world.} We provide two examples in which the verbs “throw” and “clean” in the language instructions are not in our dataset and are unseen by the model.}
   \label{real_world}
   \vspace{-0.36cm}
\end{figure}

To evaluate the generalization ability of our model in the real world, we collect some data from everyday life for testing. And the language instructions and affordance categories are unseen to the model, which poses a greater challenge. As illustrated in the \cref{real_world}, the inference results indicate that the model effectively leverages the strengths of vision-language models. By utilizing the prior world knowledge learned during the pre-training process and fine-tuning it on our dataset, the model can be easily transferred to other new scenarios in real world, achieving a complex and comprehensive understanding of semantics and spatial relationships.

\section{Conclusion}
In this paper, we introduce the task of grounding 3D object affordance based on language instructions, visual observations, and interactions. We present the first 3D affordance dataset combining images, point clouds, and language instructions, covering full-view, partial-view, and rotation-view. And inspired by cognitive science, we propose LMAffordance3D which leverages a vision-language framework to fuse 2D, 3D spatial features with semantic features, achieving part-level affordance estimation. We conduct experiments in both seen and unseen settings, and the results demonstrate the effectiveness and better generalization capabilities of our approach.

In future work, it will be interesting to explore model quantization and compression, and deploy them on robots to achieve real-time inference, combined with planning and control modules to accomplish tasks such as manipulation and operation. Moreover, since the objects in our dataset are all rigid, investigating affordance prediction for flexible objects is another promising direction for future research.

\section*{Acknowledgment}
This work was supported in part by the Joint Funds of the National Natural Science Foundation of China under Grant U24A20128, and in part by the National Nature Science Foundation of China under Grant 62173293.

{\small
\bibliographystyle{ieee_fullname}
\bibliography{main}
}

\end{document}